\documentclass[10pt,letterpaper]{article}
\usepackage[top=0.85in,left=1.5in,footskip=0.75in,marginparwidth=2in]{geometry}

\usepackage[utf8]{inputenc}
\usepackage{enumitem}

\usepackage{cite}

\usepackage{nameref,hyperref}
\usepackage{natbib}
\usepackage{amsmath}

\usepackage[right]{lineno}

\usepackage{microtype}
\DisableLigatures[f]{encoding = *, family = * }

\raggedright
\usepackage{changepage}

\usepackage[aboveskip=1pt,labelfont=bf,labelsep=period,singlelinecheck=off]{caption}

\makeatletter
\renewcommand{\@biblabel}[1]{\quad#1.}
\makeatother

\usepackage{lastpage,fancyhdr,graphicx}
\usepackage{epstopdf}
\fancyheadoffset[L]{2.25in}
\fancyfootoffset[L]{2.25in}

\usepackage{color}

\definecolor{Gray}{gray}{.25}

\usepackage{graphicx}

\usepackage{sidecap}

\usepackage{wrapfig}
\usepackage[pscoord]{eso-pic}
\usepackage[fulladjust]{marginnote}

\newcommand{\be}{\begin{eqnarray}}
\newcommand{\ee}{\end{eqnarray}}

\reversemarginpar

\begin{document}
\vspace*{0.35in}

\begin{flushleft}
{\Large
\textbf\newline{The Role of Conditional Independence in the Evolution of Intelligent Systems.}
}
\newline
\\
Jory Schossau\textsuperscript{1,2,*},
Larissa Albantakis\textsuperscript{3},
Arend Hintze\textsuperscript{4,1,2}
\\
\bigskip
\bf{1} Department of Computer Science \& Engineering, Michigan State University
  
\bf{2} BEACON Center for the Study of Evolution in Action

\bf{3} Department of Psychiatry, University of Wisconsin

\bf{4} Department of Integrative Biology, Michigan State University

\bigskip
* Corresponding author: jory@msu.edu

\end{flushleft}

\section{Abstract}
Systems are typically made from simple components regardless of their complexity. While the function of each part is easily understood, higher order functions are emergent properties and are notoriously difficult to explain. In networked systems, both digital and biological, each component receives inputs, performs a simple computation, and creates an output. When these components have multiple outputs, we intuitively assume that the outputs are causally dependent on the inputs but are themselves independent of each other given the state of their shared input~\cite{pearl2014probabilistic}. However, this intuition can be violated for components with probabilistic logic, as these typically cannot be decomposed into separate logic gates with one output each. This violation of conditional independence on the past system state is equivalent to \textit{instantaneous} interaction --- the idea is that some information between the outputs is not coming from the inputs and thus must have been created instantaneously. Here we compare evolved artificial neural systems with and without instantaneous interaction across several task environments. We show that systems without instantaneous interactions evolve faster, to higher final levels of performance, and require fewer logic components to create a densely connected cognitive machinery.


\section{Introduction}
\begin{figure*}
\centering
\includegraphics[width=5.0in]{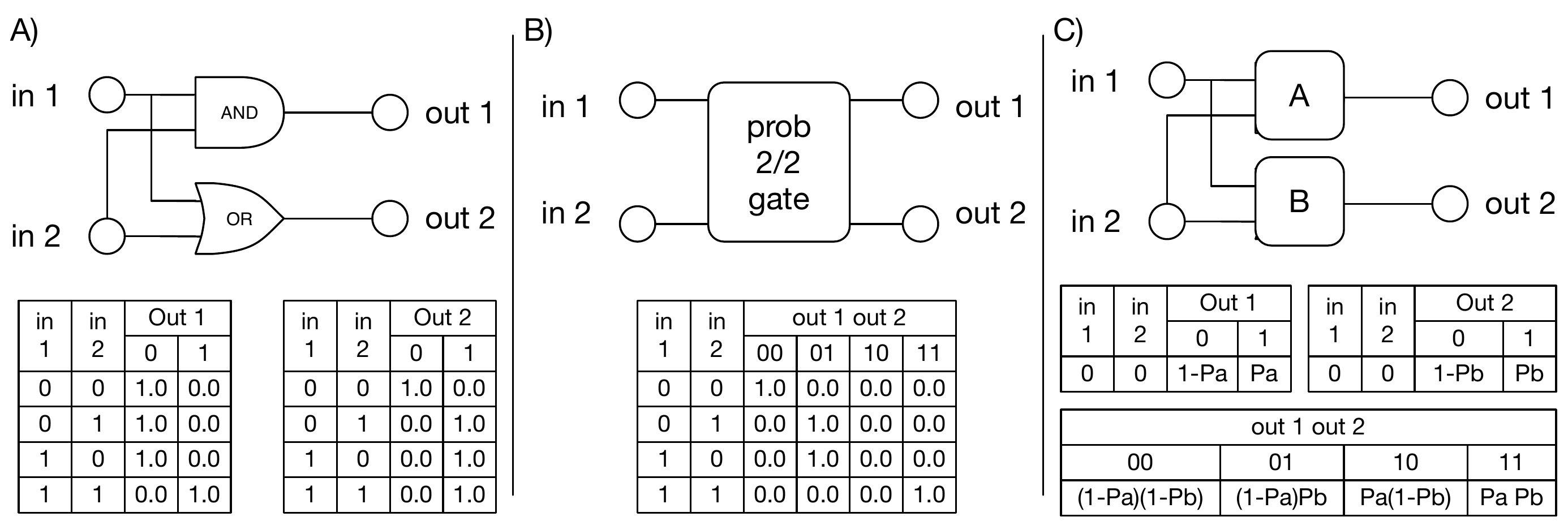} 
\caption{Examples of gate decomposition into combinations of simpler gates. Panel A shows two deterministic logic gates whose inputs are cross-wired so that both gates receive the same inputs. The tables below show their probabilities to output $0$ or $1$ respectively. These probabilistic logic boundary cases are effectively deterministic logic gates. Panel B shows a two-in-two-out logic gate that is functionally identical with the two gates depicted in Panel A. Panel C shows two probabilistic logic gates similarly connected like the deterministic gates from panel A. The logic tables below only show the case where both inputs are $0$. The lower table shows replacing both probabilistic logic gates with a single two-in-two-out probabilistic logic gate (similar to panel B) and how the new probabilities for that gate are constructed from the individual probabilities of both gates.}
\label{fig:gateComparison}
\end{figure*}

Evolvable Markov Brains are networks of deterministic and probabilistic logic gates whose function and connectivity are genetically encoded. They are a useful model to study the evolution of behavior~\cite{olson2013predator}, cognitive properties~\cite{marstaller2013evolution}, and neural-network complexity~\cite{edlund2011integrated,albantakis2014evolution}, and can also be used as classifiers~\cite{chapman2013evolution}.
At each generation of evolution within a particular task environment, networks are selected based on their fitness and the populations adapt through random genomic mutations. The genome is sequentially processed with specific sites indicating the start of a gene. An individual gene encodes one Hidden Markov Gate (HMG), which specifies connections between network elements and also determines input-output logic~\cite{marstaller2013evolution}. These HMGs are generalized logic gates that encompass conventional logic gates such as XOR or NAND, whose logic table is typically a static mapping of two inputs to a single output (Figure~\ref{fig:gateComparison}A), but allow for more than the typical two-in-one-out format and can use a probabilistic mapping between input and output states. Here, HMGs could receive up to four inputs mapped to maximally four outputs. In this way, each gene may encode an entire logic module, as opposed to only a single logic function.

Depending on the environment, deterministic or probabilistic HMGs may provide a mutually exclusive advantage for evolution. Apart from introducing randomness into the Markov Brains, probabilistic HMGs also differ from deterministic HMGs in the way they can be represented by a collection of simpler logic gates. The outputs of a deterministic HMG are necessarily conditionally independent from each other: given the input state, an output is either on (`0') or off (`1') with probability $P = 1.0$. Information about the state of other outputs is irrelevant. As a consequence, a deterministic HMG can always be decomposed into several logic gates with one output each. For example, a two-in-two-out deterministic HMG (see Figure~\ref{fig:gateComparison} panel B) can easily be decomposed into two independent two-in-one-out gates (see Figure~\ref{fig:gateComparison} panel A). This decomposition works similarly for larger gates with more inputs and more outputs, requiring one logic gate per output.

Probabilistic HMGs, on the other hand, are not generally decomposable into separate logic functions for each output. In the case of a two-in-two-out probabilistic HMG, a probability table (for a detailed explanation see~\cite{marstaller2013evolution}) maps all four possible input states to all four possible output states. 
Let us for simplification purposes just consider the case where both inputs for such a gate are $0$. Then, four probabilities ($P_{00}$, $P_{01}$, $P_{10}$, $P_{11}$) determine the probability for each of the four possible output states to occur given input state $00$. $P_{00}$ defines the probability that both outputs are 0, and $P_{11}$ the probability that both outputs are 1, and so forth. Observe that all four probabilities must sum to $1.0$: 
\be
P_{00}+P_{01}+P_{10}+P_{11}=1.0
\label{equ:summation}
\ee
Except for the above requirement, the individual probabilities $P_{00}$ to $P_{11}$ evolve independently for default probabilistic HMGs in Markov Brains.

Let us now try to use two probabilistic two-in-one-out gates (see Figure~\ref{fig:gateComparison} panel C) to achieve the same functionality as a two-in-two-out HMG. $P_a = P(A=1|I)$ denotes the probability of the first gate (`A') to have an output of $1$ for a given input state $I$. Consequently, the probability for A to have an output of $0$ is $1.0-P_a$. Synonymously, for the second gate (`B')  $P_b = P(B=1|I)$. The joint input-output function of the two individual gates A and B is the following: 
\be
P_{00}=(1.0-P_{a})(1.0-P_{b})\\
P_{01}=(1.0-P_{a})P_{b}\\
P_{10}=P_{a}(1.0-P_{b})\\
P_{11}=P_{a}P_{b}
\label{equ:requirements}
\ee
Given eqs. 2-\ref{equ:requirements}, the summation rule (eq.~\ref{equ:summation}) is met. In addition, the following dependency between probabilities holds:
\be
P_{00}P_{11}=P_{01}P_{10}
\label{equ:violation}    
\ee
It is easy to see that probabilistic HMGs with independently evolved probabilities $P_{00}$ to $P_{11}$ may violate equation~\ref{equ:violation}. As a result, probabilistic HMGs typically cannot be decomposed. Decomposition of a probabilistic two-in-two-out HMG is only possible if:
\be
P(\text{out}_1|I, \text{out}_2) = P(\text{out}_1|I)\\
P(\text{out}_2|I, \text{out}_1) = P(\text{out}_2|I)
\label{equ:condIndependence}    
\ee
for any input state $I$, which means that the two outputs must be conditionally independent of each other given all possible $I$. Under these conditions, the HMG's probabilities can be expressed according to eqs. 2-5, with $P_a = P(\text{out}_1=1|I)$, the marginal probability of $\text{out}_1 = 1$ given $I$, and $P_b = P(\text{out}_2=1|I)$, the marginal probability of $\text{out}_2 = 1$ given $I$. The same principle can be applied to HMGs with multiple conditionally independent outputs.

An example probability distribution for $P_{00}$ to $P_{11}$ that violates conditional independence is $\text{PD} = P(\text{out}_1, \text{out}_2|I)=$ (0, 0.5, 0.5, 0), whereas $\text{PD}^*=$ (0.25, 0.25, 0.25, 0.25) conforms with eqs. 1-8. Note that the marginal probabilities $P(\text{out}_1=1|I)=P(\text{out}_2=1|I)=0.25$ are the same in both cases. In fact, there are infinitely many probability distributions with the same marginal probabilities, but only $\text{PD}^*$ fulfills conditional independence given the input state. By contrast, PD contains the additional constraint that $P(\text{out}_1=1|I,\text{out}_2=1) = 0$ and vice versa (cf.~\cite{james2016multivariate} for more intricate examples of hidden dependencies in probability distributions). Making the temporal order explicit, the probabilities in $\text{PD}^*$ only depend on the input $I$ to the HMG at timestep $t-1$, before the update. PD, however, also requires instantaneous interaction between the outputs at time $t$. 

This example demonstrates that, in Markov Brains with probabilistic HMGs, the output state of an element at time $t$ may depend on information that is not available at $t-1$. Such instantaneous interactions have implications with respect to the causal structure of these Markov Brains, as they violate the postulate that causes must precede their effects. In addition, instantaneous interactions defy the notion of elementary causal components. Without conditional independence given the input state, the interactions between elements in a probabilistic Markov Brain cannot be represented as a directed acyclic \textit{causal} graph~\cite{pearl2014probabilistic}. This prohibits analyzing the causal composition of these Markov Brains, which means, for example, that the theoretical framework of integrated information theory (IIT)~\cite{oizumi2014IIT, albantakis2015intrCausation}, which assesses how sets of elements within a system causally constrain each other, cannot be applied to these Markov Brains. 
In short, while Markov Brains with general, probabilistic HMGs may be useful tools for artificial evolution experiments, the networks of elements they encode cannot generally be interpreted as a network of causally interacting components. 

While instantaneous interaction may be a curious phenomenon in evolvable Markov Brains, the question remains whether the potential gain in computational power through such instantaneous interactions has any effect on the evolution and functionality of Markov Brains. To explore this question, we implemented a \textit{decomposable} version of the evolvable probabilistic HMGs (up to four-in-four-out) with conditionally independent outputs $\{\text{out}_1, \dots, \text{out}_N\}$, such that:
\be
P(\text{out}_i|I, \{\text{out}_{1:N \setminus i}\}) = P(\text{out}_i|I),
\label{equ:condIndependenceGen}    
\ee
for all input states $I$ and all outputs $\{\text{out}_1, \dots, \text{out}_N\}$, where $N$ is the number of outputs. These decomposable HMGs comply with an extended version of eqs. 2-5:
\be
P(\text{out}_{1:N} = O|I) = \prod_{i=1:N} P(\text{out}_i = O_i|I)
\label{equ:decompLogicTable}    
\ee
for all output states $O$ and all input states $I$, and thus guarantee Markov Brains with causally interpretable neural networks. In the following, we will compare the evolution of Markov Brains generated by probabilistic HMGs against Markov Brains generated by decomposable probabilistic HMGs in several task environments. 

We will show that instantaneous interaction hampers evolutionary adaptation. At the same time we show that systems evolved from components without instantaneous interaction adapt their cognitive machinery. Contrary to what might be expected given the potential gain in computational power through instantaneous interactions, Markov Brains with decomposable HMGs required fewer gates to create a more densely connected network with better task performance. We conjecture that instantaneous interactions provide no computational advantage for agents evolving in the tested sensory-motor task environments.

\section{Results}
Probabilistic HMGs in Markov Brains can independently evolve the probabilities that determine their outputs, which can lead to instantaneous interactions, as outlined above. Here, we thus introduce another type of probabilistic HMG we call \textit{decomposable HMG}.
These gates encode their size and connectivity in the same way as probabilistic HMGs. Nevertheless, the probability tables of decomposable HMGs are restricted according to eqs. 9 and 10, and thus must be encoded in a different way than those of general probabilistic HMGs. 

Probabilities are encoded as one number per one locus. In the non-decomposable HMG these probabilities are transcribed directly from the genome to fill the probability table, after which they are appropriately normalized given the table dimensions. For decomposable HMGs we ensure decomposability by transcribing only the marginal probabilities (eq.~\ref{equ:condIndependenceGen}). The probability matrix is then created from these marginal probabilities $P(\text{out}_i|I)$ by using eq.~\ref{equ:decompLogicTable}. This allows us to evolve Markov Brains with either or both types of gates, conventional probabilistic and decomposable, or in other words those which have instantaneous interaction and those which do not. The difference in rate of adaptation, and differences in the solutions evolved, will highlight the effect of instantaneous interaction on evolution.

Differences between Markov Brains with and without instantaneous interactions may depend on the environments in which the virtual organisms were evolved. Some environments might benefit from instantaneous interactions, while others might favor conditional independence. For this reason, three different environments were tested:
\begin{description}[wide, labelwidth=!, labelindent=0pt]
\item[Temporal Spatial Integration]\hfill \break
In this environment~\cite{marstaller2013evolution,albantakis2014evolution} the agent can move laterally left and right (one binary effector each) and is equipped with a set of upwards facing sensors, two on the left and two on the right (one binary sensor each) with a gap of 2 block subunits between them, and 8 hidden binary elements for storing information. Small and large blocks are falling toward the agent one at a time, and activate the sensors of the agent when above them regardless of distance. The blocks fall in different directions, and the block sizes and sensors are arranged in such a way that the blocks need to be observed over several updates in order to be distinguished. Small blocks must be caught while large blocks must be avoided. 
\break

\item[Foraging and Spatial Reasoning]\hfill \break
Agents are placed at a designated home area in a two-dimensional environment and must first discover and then obtain food. Once they obtained food, they must move it to the home location, after which more food must be collected. Early in an agent's life food appears nearby, but the circular perimeter onto which food is randomly placed increases in diameter with each successful collection, moving the food successively farther away from home. The sensors for this agent provide a very coarse representation of the environment and are only accurate for nearby objects. Inputs include sensor signals for food and home locations: on, facing, and near, as well as angle to food with perception of angle discretized to 45 degrees. Consequently, food and home locations are not reliably observable, and the agent must navigate heuristically in the absence of this sensory information. Outputs include turning left or right and moving forward. 
\break

\item[Associative Memory]\hfill \break
This two-dimensional environment~\cite{grabowski2010early,grabowskibuilding} presents the agent with a path of rewards, surrounded by a field of poison. The agent receives 4 sensor inputs about whether its current location is on path, poison, or one of the cues. In addition, it receives cues about upcoming turns in the path. These cues, their turn associations, and the path itself are all randomly generated with each visit to the environment. The agent may affect 2 outputs which encode 4 actions: nothing, turn left one unit, turn right one unit, and move forward one unit. In addition, the agent has 8 hidden binary elements for memory. 
The agent must explore the environment and learn which of the two symbols indicates right and which one indicates left. The agent must then remember those associations and use that knowledge to navigate the path properly.

\end{description}

For each experimental condition, we evolved 120 independent populations of 100 organisms until their performance plateaued. A plateau was reached near generation 10,000 in the temporal spatial integration environment, generation 5,000 in the associative learning environment, and generation 3,000 in the foraging environment. After evolution the line of descent was reconstructed~\cite{lenski2003evolutionary}, that is the path of inheritance from a random organism in the final population to its ancestor in the initial population. All sweeping mutations are observable in the line of descent.

\begin{figure}
\centering
\includegraphics[height=2.0in]{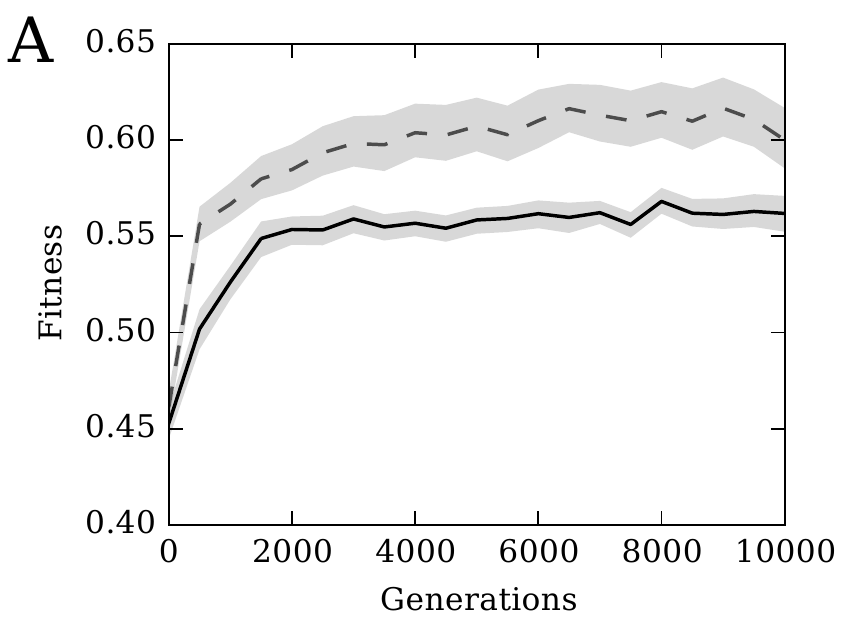}
\includegraphics[height=2.0in]{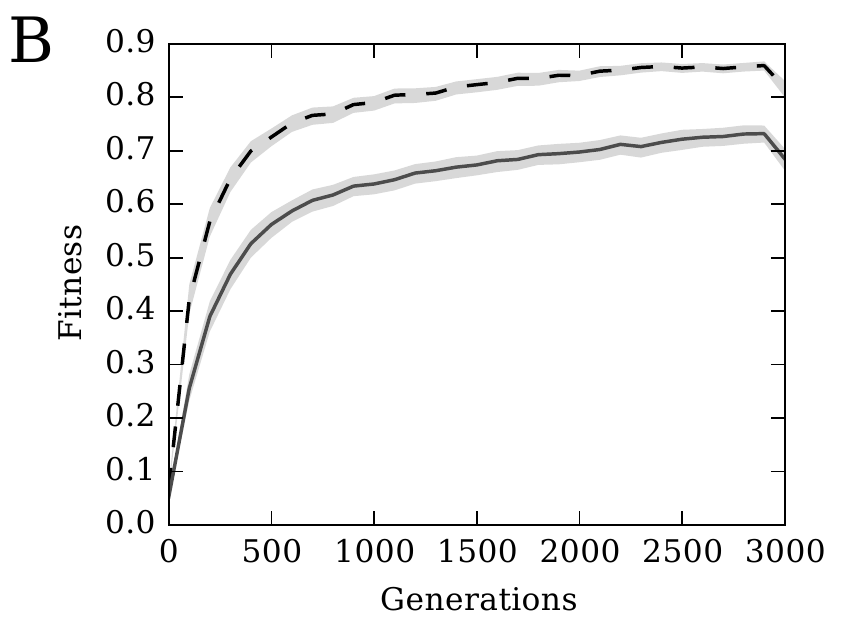} 
\caption{Average fitness of organisms on the line of descent in the spatial temporal (A) and foraging (B) environments. The solid line represents average performance of agents restricted to conventional probabilistic HMGs (with instantaneous interactions), dashed lines represents average performance of agents restricted to decomposable HMGs (without instantaneous interactions). The y axes are normalized to show the fraction of maximally attainable fitness in each environment. The gray shadow indicates the bootstrapped 95\% confidence interval of the mean.}
\label{fig:performanceSTIandForage}
\end{figure}

We find that in the foraging and temporal spatial integration environments agents restricted to decomposable gates adapt much more quickly and also achieve a better final performance (see Figure~\ref{fig:performanceSTIandForage}).

\begin{figure}
\centering
\includegraphics[height=2.0in]{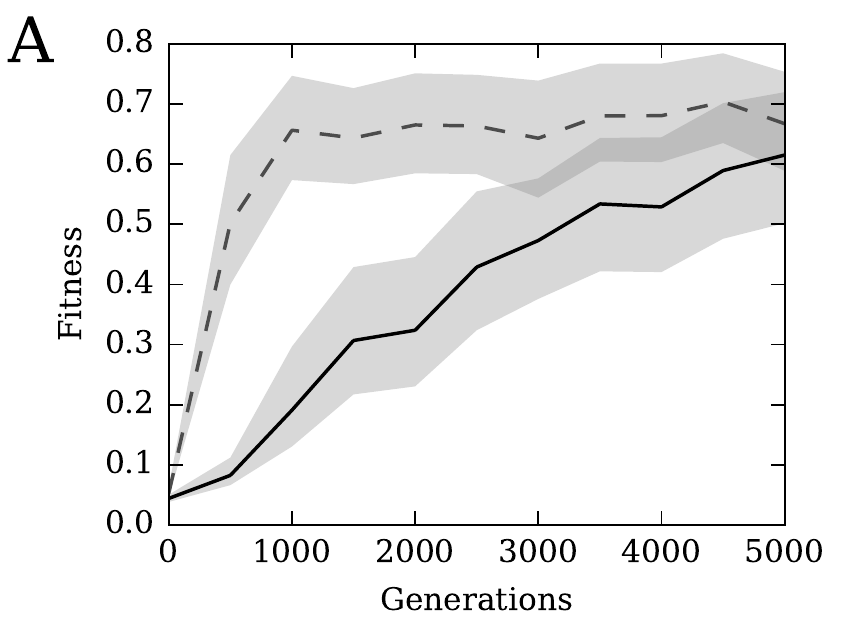} \includegraphics[height=2.0in]{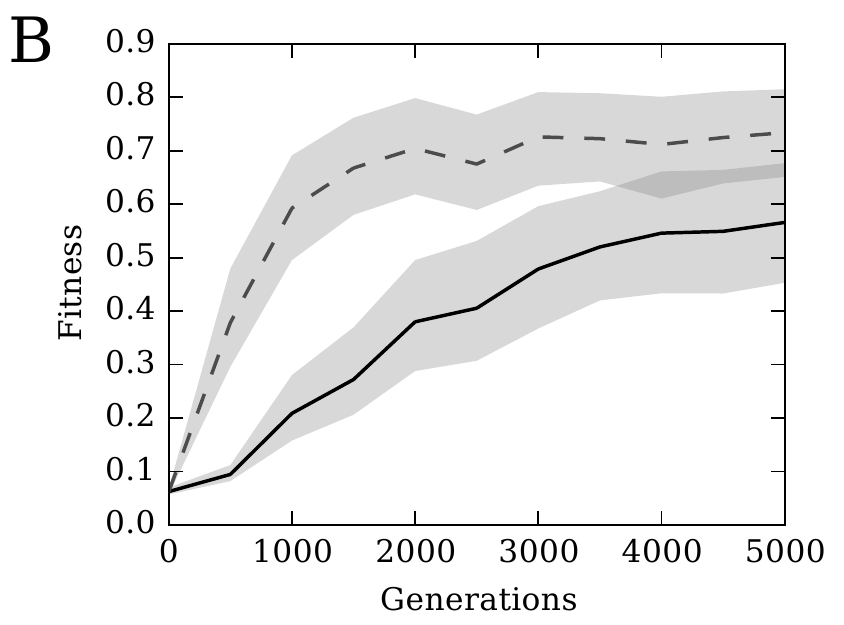} \includegraphics[height=2.0in]{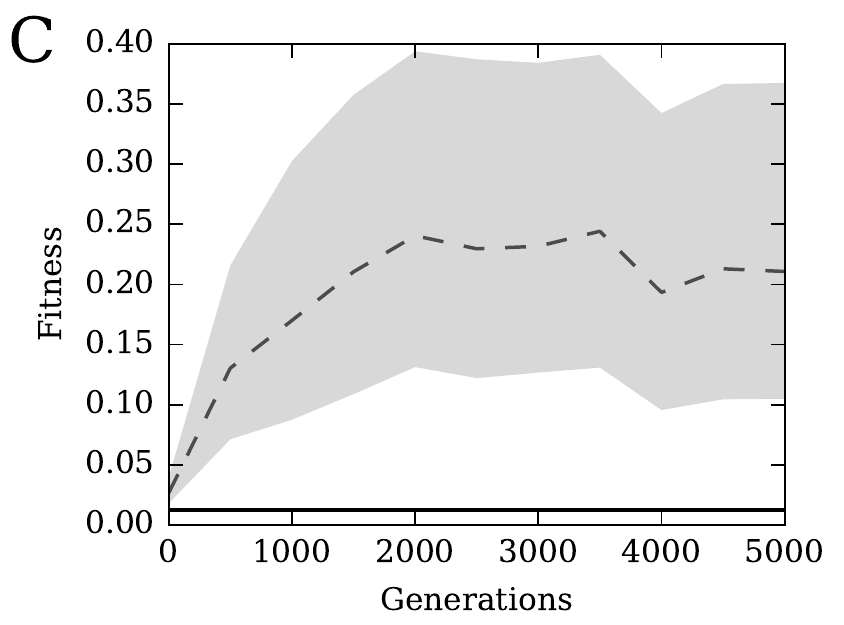} 
\caption{Average performance of agents evolved in the association environment with increasing levels of error punishment. Performance of agents evolved while restricted to decomposable HMGs is shown as a dashed line. Performance of agents evolved while restricted to conventional probabilistic HMGs is shown as a solid line. The gray shadow indicates the bootstrapped 95\% confidence interval of the mean. Panel A shows the results for agents receiving zero punishment for path-following errors. Panel B shows the results for punishment of $0.05$, meaning $0.05$ was subtracted from the cumulative agent score every time it wandered off the path. Panel C shows the results for agents evolved with punishment $0.1$, again subtracted from their score when wandering off the path. The y axes were normalized to show relative performance, with $1.0$ being the maximally attainable fitness in each environment assuming ideal behavior.}
\label{fig:performanceAssociationTask}
\end{figure}

The associative memory environment has a tunable punishment parameter for the cost of wandering off the correct path. The relative difference between the reward for following the path and the punishment for straying from the path greatly influences evolvability of successful strategies. If wandering off the path is very costly, then the agent is severely limited in its freedom to make mistakes and explore if it is to maintain its status as a viable organism. This harsh limitation on mistake-making hampers evolution. We find that when punishment for path deviation is relatively small or zero, decomposable HMGs provide a clear evolutionary advantage. When punishment is high, agents restricted to conventional probabilistic HMGs fail to evolve at all, while agents restricted to decomposable HMGs still evolve functional Markov Brains (see Figure~\ref{fig:performanceAssociationTask}). 

\begin{figure}
\centering
\includegraphics[height=2.0in]{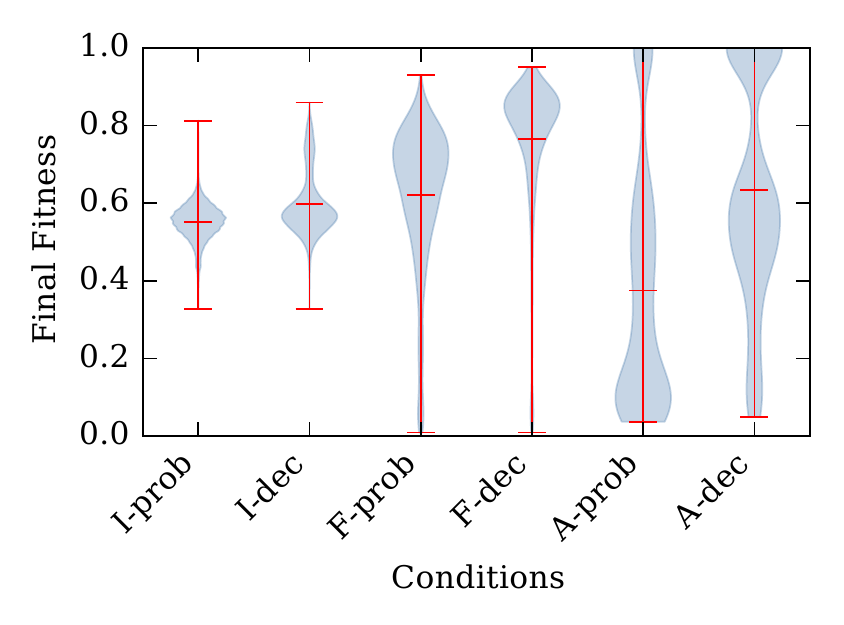} 
\caption{Distribution of performances at the end of evolution for each of two conditions in each of three environments. At the end of evolution, the 200\textsuperscript{th}-from-last agent of each independent line of descent is collected to create these distributions. This near-the-end data slicing is necessary to eliminate noise toward the end of the line of descent caused by mutated extant agents not yet pruned from the line of descent by selection.
The three environments are represented on the x-axis as I (temporal spatial integration task), F (foraging task), and A (association task with a punishment of $0.05$). In each environment (I, F, and A) agents were allowed to either use conventional probabilistic HMGs (labeled as ``prob'') or decomposable HMGs (labeled as ``dec''). Red dashes indicate the mean and extrema. Gray violin plots show the distributions of normalized fitness for all 120 replicates per experimental condition. Fitness was normalized such that maximal theoretically attainable fitness is represented as $1.0$. For each environment of Integration, Foraging, and Association, the conditions under which evolution was limited to decomposable gates produced significantly better adapted agents. Significance was tested using the Mann-Whitney U test, 
with $p < 0.05$ for each environment ($p=0.0$, $p=0.0$, $p<2.2 \times 10^{-112}$).}
\label{fig:scoreDistribution}
\end{figure}

As shown in Figure~\ref{fig:scoreDistribution}, it is not only the rate of adaptation that is higher for decomposable gates, but also the mean final achieved performance in all three environments, even though the extrema and distributions vary greatly by environment.

\begin{figure}
\centering
\includegraphics[height=2.0in]{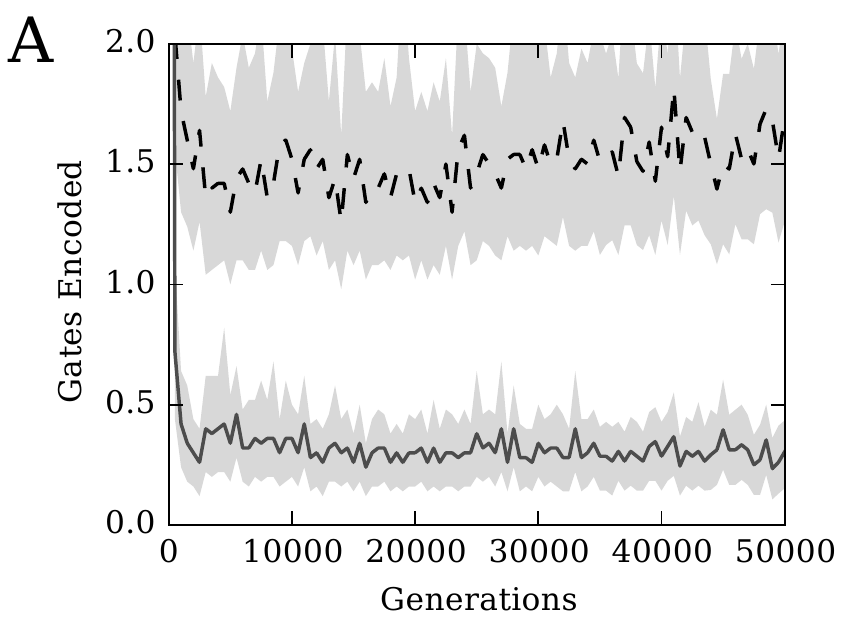} \includegraphics[height=2.0in]{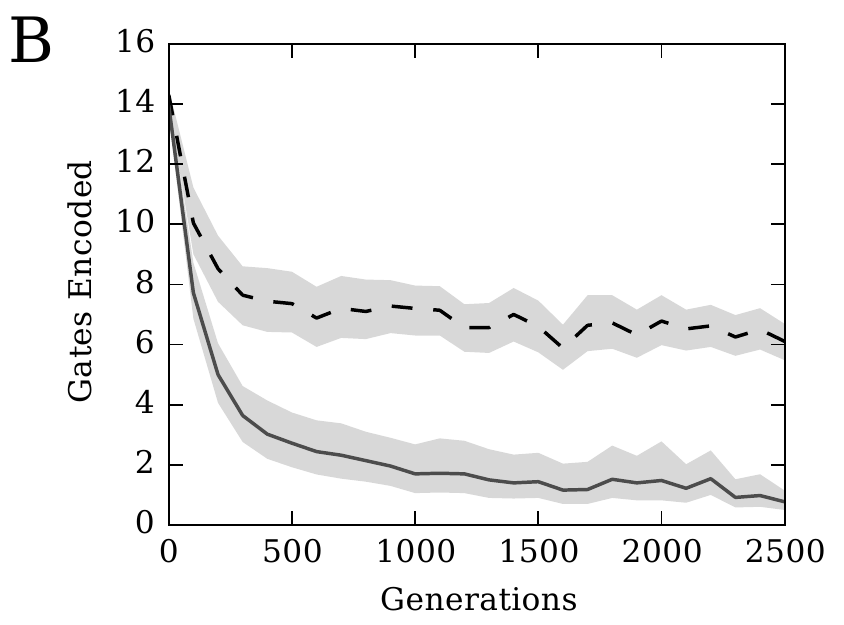} \includegraphics[height=2.0in]{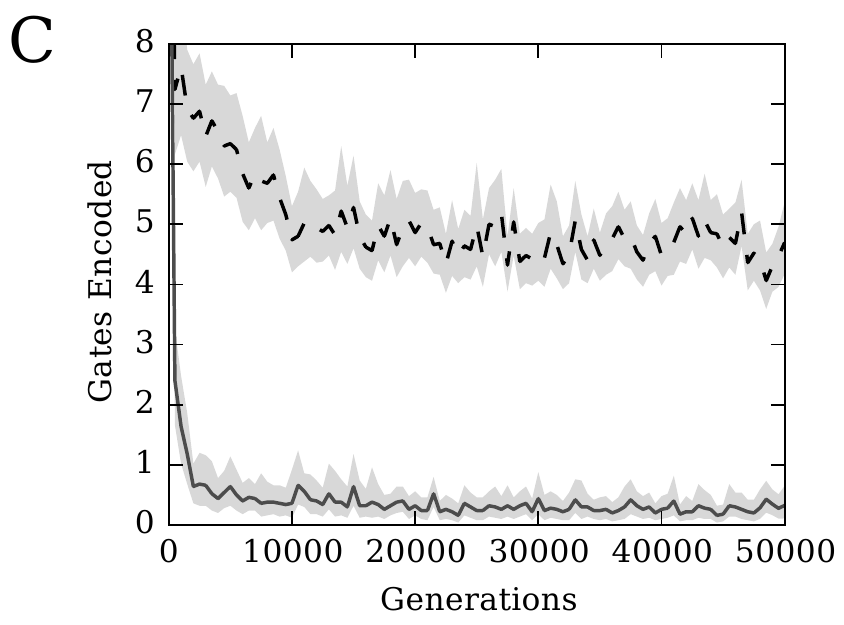} 
\caption{Average number of gates evolved along the line of descent in a competition experiment executed in three different environments. Panel A shows the results from the temporal spatial integration environment. Panel B shows the results from the foraging environment. Panel C shows the results from the association environment with a punishment cost of $0.05$. The average number of preferentially selected decomposable HMGs over generations is shown as a dashed line, whereas the solid line represents the average number of preferentially selected conventional probabilistic HMGs.
The gray shadow indicates a 95\% confidence interval.}
\label{fig:competition}
\end{figure}

In the above simulations, Markov Brains were restricted to one type of HMG, either probabilistic or decomposable and seeded with 15 gates at the start of evolution. In a second set of experiments we allowed Markov Brains to evolve both types of gates in order to assess if there is preferential selection of one gate over the other. If either of the gate types confers more of an advantage it should be selected more often than the other gate type. The null hypothesis suggests both gates confer the same benefit and would be under equal selection. While it took much more computational power to simulate these populations longer, we wanted to investigate if there might be oscillations in preferential selection for gate type. In all tested environments we find that decomposable HMGs are used more often than conventional probabilistic HMGs (see Figure~\ref{fig:competition}). To exclude any variation resulting from the initial evolvability differences between the gates, agents are seeded with 15 HMGs of each type. We find that selection quickly reduces these, but decomposable HMGs are kept more often than probabilistic HMGs. Observe that different tasks require brains of different sizes, but the effect of the dominating decomposable gate type is independent of this phenomenon (see Figure~\ref{fig:competition}). 50 independent populations of 100 agents each were evolved in each environment for these competitions.

So far, our results suggest that decomposable HMGs have a clear advantage when it comes to evolutionary adaptation of Markov Brains. Furthermore, these results suggest that systems without instantaneous interactions evolve faster, and select against instantaneous interaction when possible. Instantaneous interaction allows components to have outputs which share information beyond that of their inputs. As such, the question was if instantaneous interactions provide an advantage to computation. Our results suggest the opposite. The question is now if systems which cannot have instantaneous interactions compensate for that loss. It may be that decomposed components are more adaptable in a way that compensates for the missing information. To investigate this issue we also assessed the effect of gate type on the structure of cognitive machinery. Observe that the different kinds of gates only differ in how their probabilities are encoded and do not differ in how connections are made or how mutations affect their connection or abundance. While there are many ways to measure these networks~\cite{McCabe1976} we test evolved brains for number of gates and graph measures used in similar work: gamma index (connectivity), and brain diameter (for a more detailed explanation see \cite{schossau2015information}).The first measure is simply the number of gates present in the phenotype. The Gamma index measures the density of connections relative to all possible connection which could have been made. A higher gamma index indicates a more connected Markov Brain. 

\begin{figure}
\centering
\includegraphics[height=2.0in]{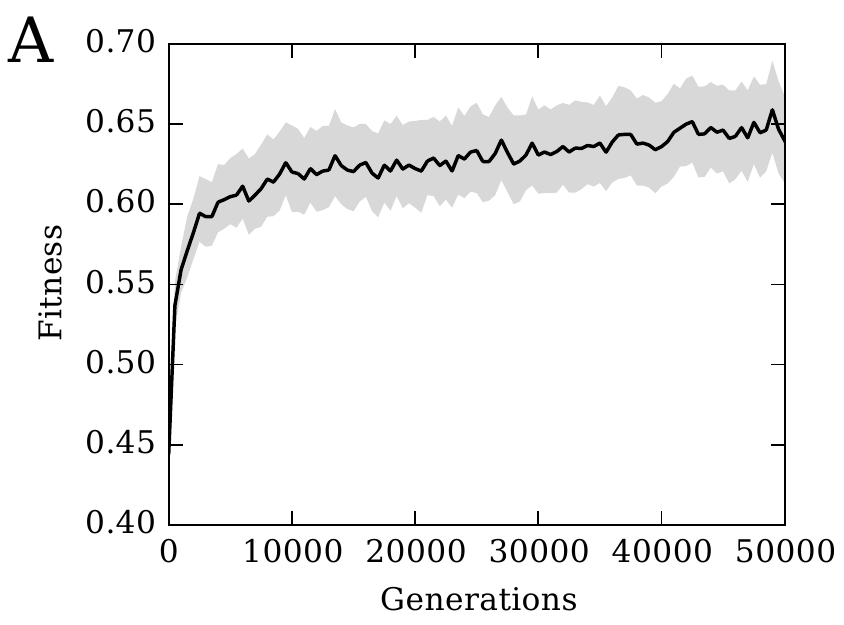} \includegraphics[height=2.0in]{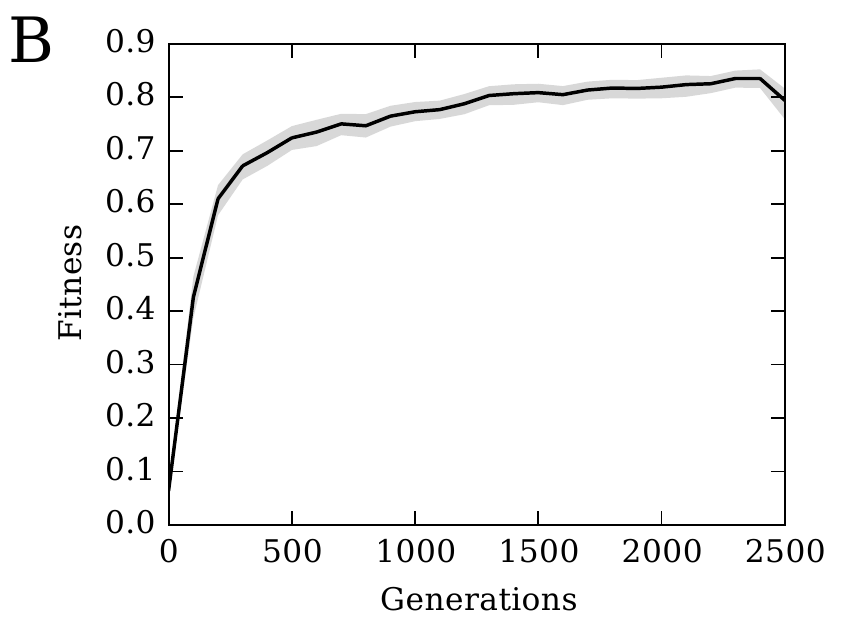} \includegraphics[height=2.0in]{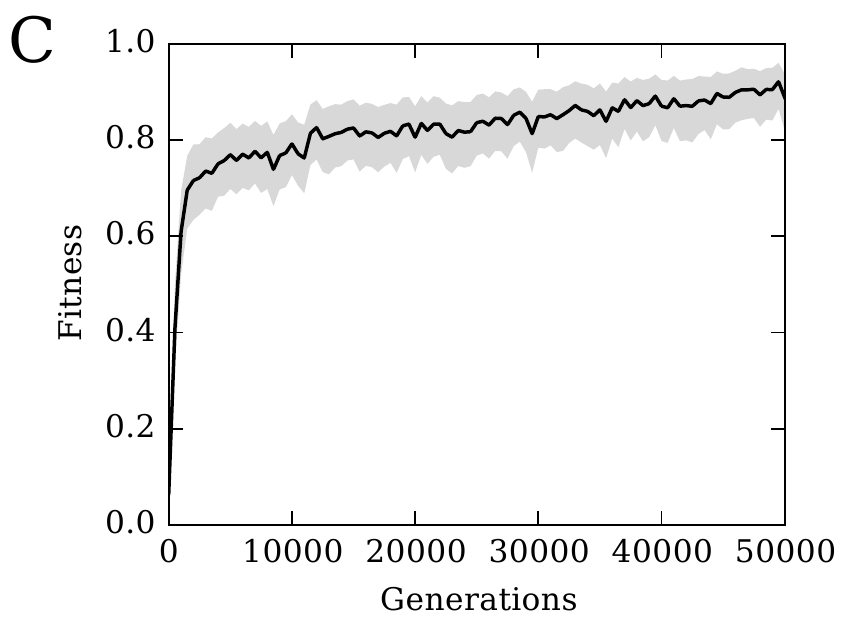} 
\caption{Average fitness along the line of descent in each environment while both non-decomposable and decomposable gate types were evolvable: Panel A shows line of descent fitness for the temporal spatial integration environment. Panel B shows line of descent fitness for the foraging environment. Panel C shows line of descent fitness for the association environment. 
The gray shadow indicates a 95\% confidence interval.}
\label{fig:competitionFitnesses}
\end{figure}

The brain diameter is the length of the shortest path between the furthest nodes in the network. Diameter is assessed by computing all shortest connections between all components. The longest of these shortest connections is the brain diameter. The larger the brain diameter is the more steps it takes for information to traverse the Markov Brain, and the more computations are possible. 

Probably the most interesting result is that the total number of gates was lower when only decomposable gates were allowed, at least when agents were evolved in the foraging and in the association environment (see Figure~\ref{fig:neuroProperties} panel C). When agents were evolved in the temporal spatial integration environments, we find a slightly higher number of decomposable gates. This suggests that the subsequent effects on density and diameter cannot simply be explained by a higher number of gates. We find that the density and diameter of the evolved Markov Brains are higher for those experiments where agents were restricted to the evolution of decomposable gates (see Figure~\ref{fig:neuroProperties} panel B and C). This suggests agents with decomposable gates evolve to use fewer gates, and that these fewer gates are more tightly connected. Additionally, the total diameter of the network becomes longer. This result suggests a possible contradiction to the idea that a system needs more decomposable gates to compensate for the lack of instantaneous interactions. Secondly, it suggests that if the loss of instantaneous interactions is part of a trade-off, then it is offset by an increase in connectivity and brain diameter, though it may only be that these populations evolve faster and thus arrive at better-connected high-fitness networks sooner. The increase in brain diameter suggests that more computations are occurring sequentially in the brain. However, we have no intuition what this could mean for the computations that happen in the Markov Brain.

While the mixed gate conditions favored evolution of decomposable gates, this could have come at the cost of lower achieved fitness than in the homogeneous gate conditions. However, an examination of the line of descent fitness show no degradation in evolvability. This suggests that while the path for a slower and more complex evolutionary trajectory with non-decomposable gates existed within the search space, evolution preferentially selected against that path for one with more evolvability.

\begin{figure}
\centering
\includegraphics[height=2.0in]{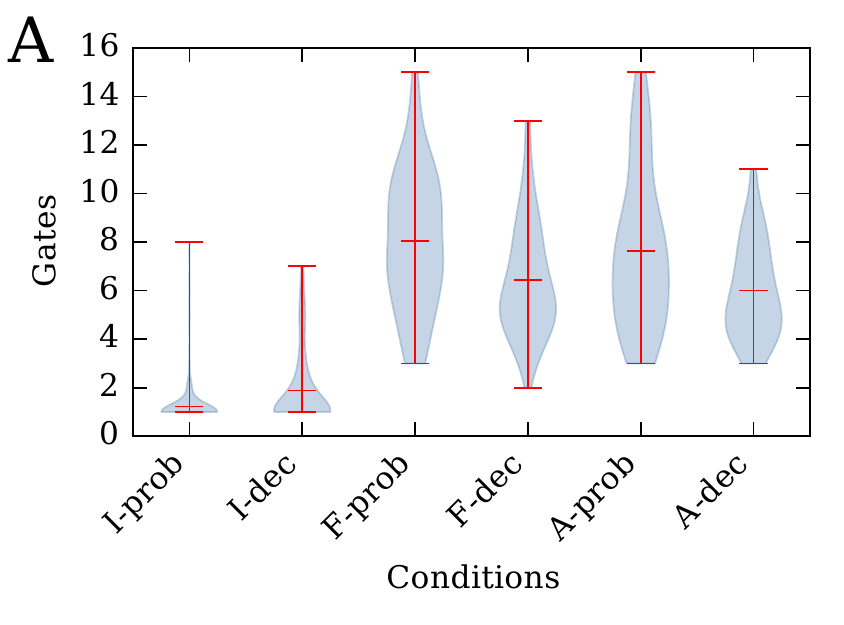}
\includegraphics[height=2.0in]{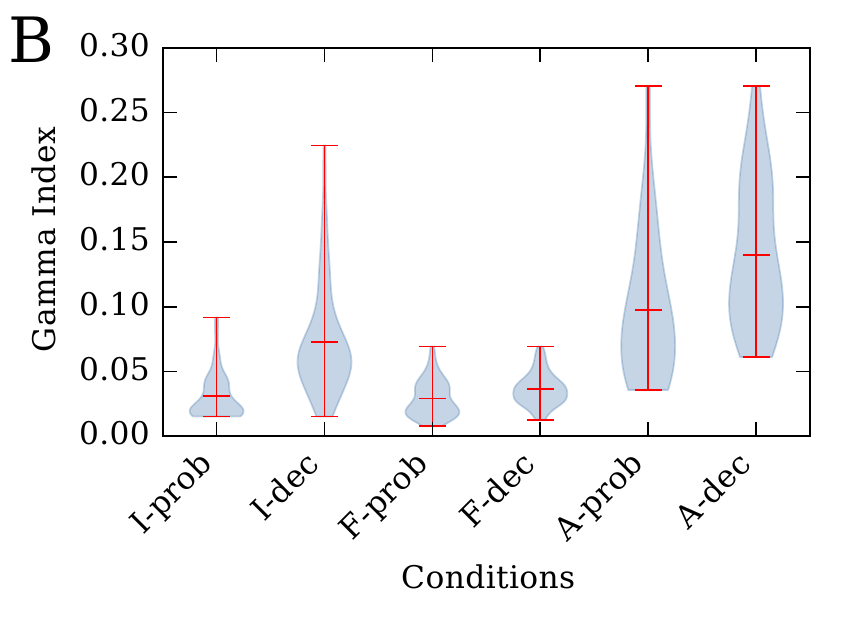} 
\includegraphics[height=2.0in]{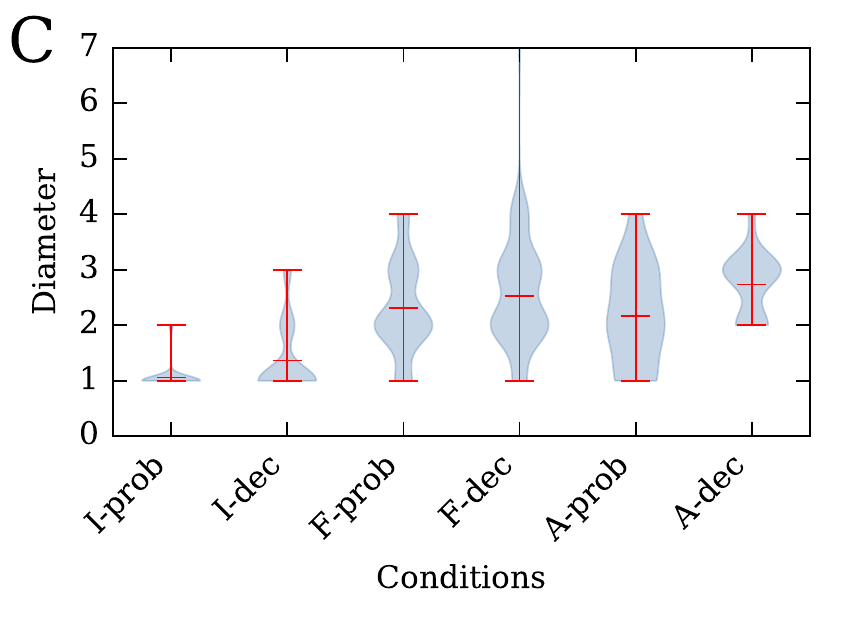} 
\end{figure}
\begin{figure}
\caption{Distributions of different properties of Markov Brains at the end of evolution in three environments and two conditions for each environment. The three environments are represented on the x-axes as I (temporal spatial integration task), F (foraging task), and A (association task with a punishment of $0.05$). In each environment (I, F, and A) agents were allowed to either use conventional probabilistic gates (labeled as ''prob") or decomposable gates (labeled as ''dec"). Panel A shows the number of gates, Panel B shows the gamma index (that is the connection density), and Panel C shows the diameter (longest of the shortest connections). Each panel shows the distributions as violin plots in gray, and the mean as well as the extreme points as red dashes. For each environment of Integration, Foraging, and Association, and for both properties of Gamma Index and Diameter, the conditions under which evolution was limited to decomposable gates produced significantly more connected and larger diameter Markov Brains. Significance was tested using the Mann-Whitney U test, with $p < 0.05$. For Gamma Index, the difference between gate restricted evolution within environments produced $p$ values of $p=4.9 \times 10^{-23}$, $p=6.5 \times 10^{-6}$, $p=1.3 \times 10^{-3}$ respectively. For Diameter, the difference between gate restricted evolution within environments produced $p$ values of $p=2.2 \times 10^{-6}$, $p=3.6 \times 10^{-2}$, $p=4.2 \times 10^{-3}$ respectively.}
\label{fig:neuroProperties}
\end{figure}

\section{Discussion}


Evolvable Markov Brains are a useful modeling framework for studying evolution~\cite{olson2013predator} and complex cognitive systems ~\cite{edlund2011integrated,marstaller2013evolution,albantakis2014evolution}, but also provide a powerful alternative to conventional machine learning approaches due to their capacity for solving classification tasks ~\cite{chapman2013evolution}. Their connectivity and logic structure is determined by genes which encode HMGs that map inputs to outputs via a logic table. Probabilistic HMGs yield the most general logic tables, as each entry is independently determined by one locus in the genome with the only restriction that each row has to sum to 1. As a consequence, probabilistic HMGs allow for instantaneous interactions, shared information between outputs. This information sharing may contain information useful to the system, but prohibits interpreting Markov Brains with probabilistic HMGs as a network of elementary causally interacting components~\cite{oizumi2014IIT,pearl2014probabilistic, albantakis2015intrCausation}. While this may not impact machine learning applications of Markov Brains, it may limit their use as a surrogate for biological systems and aggravates analyzing their causal structure. To date, the possible role of these instantaneous interactions in cognitive systems remains unclear.

Here we investigated the impact of instantaneous interactions on cognitive systems by comparing the evolution of two different types of Markov Brains. Agents were evolved in three different environments and were restricted to either conventional probabilistic HMGs or \textit{decomposable} HMGs which did not allow for instantaneous interactions (eq. 9 and 10) but were otherwise identical to conventional probabilistic logic gates.

The decomposable HMGs introduced here are a special case of probabilistic HMGs, with the additional constraint that their outputs are conditionally independent given the past system state. A priori, probabilistic HMGs thus have more computational potential. Nevertheless, we found that decomposable HMGs without instantaneous interactions not only allow faster adaptation of Markov Brains, but also promote improved final agent performance. In populations which could evolve the use of both probabilistic and decomposable HMGs, we found preferential selection for decomposable gates in all three tested environments. Lastly, we found that populations restricted to decomposable HMGs evolved to use fewer gates, have a larger diameter, and a higher density of connectivity. 

These findings suggest that instantaneous interactions hamper the evolution of cognitive systems rather than providing computational advantages. In fact, Markov Brains with decomposable HMGs evolved to higher fitness levels using a similar or even fewer number of gates than Markov Brains with probabilistic HMGs. This indicates that Markov Brains with decomposable HMGs had no need to compensate for a lack in computational power. To the contrary, evolution seems to exploit the conditional independence property of decomposable HMGs in these systems and to avoid instantaneous interactions. This is suggested by the finding that Markov Brains with decomposable HMGs are more densely connected, which means that individual decomposable HMGs evolved on average more outputs than individual probabilistic HMGs. Conditional independence thus seems to facilitate packing more input-output relations into a single HMG. In Markov Brains with probabilistic HMGs more gates with fewer average outputs may be required specifically to avoid instantaneous interactions.

Further investigation should provide more insight into the question why systems with elementary causal components that do not allow instantaneous interactions evolve faster. After all, an instantaneous interaction contains information seemingly ``from nothing'' as it is created between outputs of logic units and is not solely caused by inputs. Based on our simulations, several factors may contribute to explain why evolution prefers decomposable gates:

\begin{description}[wide=0\parindent]
\item[Search Space]
Evolutionary search for a population using probabilistic gates must traverse a larger state space, since probabilistic gates include the subset of gates which are decomposable. When limiting evolution to explore only decomposable gates, the search space becomes much smaller and thus easier to explore. 
\item[Epistatic Interactions]
In decomposable gates, mutations may have a different functional phenotypic effect than those affecting conventional probabilistic gates. This may allow transversing the evolutionary search space in greater leaps. 
\item[Robustness of Functions]
Related to the above, conditional independence between the outputs of a decomposable HMG may lead to a more robust encoding of beneficial input-output relations. 
\item[Determinism with respect to Inputs]
All tested evolution environments required the agents to react to sensor stimuli in order to achieve high fitness. While a certain amount of intrinsic indeterminism (noise) may be useful to trigger exploratory behavior, instantaneous interactions between outputs that are not explained by past inputs might simply provide no fitness advantage in sensory-motor cognitive tasks. Decomposable HMGs can be viewed as a noisy version of deterministic HMGs. Any additional freedom in the logic table of probabilistic HMGs may be superfluous or even detrimental. 
\end{description}

Our work has several practical implications for the use of Markov Brains across disciplines. For machine learning applications, where the interest lies primarily in the speed of evolution and final fitness, our results suggest that decomposable HMGs may improve performance considerably compared to general probabilistic HMGs. As models for cognitive systems, using decomposable HMGs instead of general probabilistic HMGs has the additional benefit that decomposable gates lead to causally interpretable Markov Brains. This allows analyzing the causal composition of the resulting Markov Brains, for example, within the framework of Integrated Information Theory~\cite{oizumi2014IIT, albantakis2014evolution}.

Finally, whether biology and fundamentally physics allow for true instantaneous interactions in nature is an open question. Typical accounts of causation require that causes precede their effects. Yet, missing variables and coarse-grained measurements may lead to system models with apparent instantaneous interactions between variables~\cite{james2016multivariate}. Taken together, our results suggest that there is no apparent reason to include instantaneous interactions in Markov Brains.

\section*{Acknowledgements}
L.A. receives funding from the Templeton World Charities Foundation (Grant \#TWCF0196).

\nolinenumbers

\end{document}